\documentclass[runningheads]{llncs}
\usepackage[T1]{fontenc}
\usepackage{amsmath}
\usepackage{graphicx}
\usepackage{algorithm}
\usepackage{algorithmicx}
\usepackage{algpseudocode}
\begin{document}
\title{LangSAT: A Novel Framework Combining NLP and Reinforcement Learning for SAT Solving}
%
%
%
\author{Muyu Pan\inst{1},  
Matthew Walter\inst{1},
Dheeraj Kodakandla\inst{1}
\and
Mahfuza Farooque
}
\authorrunning{F. Author et al.}
%
\institute{
Pennsylvania State University, University Park, PA  16802, USA, 
\email{\{mfp5696, mff5187,mdw5726,djk6439\}@psu.edu}}
\maketitle              
\begin{abstract}
Our work presents a novel reinforcement learning (RL) based framework to optimize heuristic selection within the conflict-driven clause learning (CDCL) process, improving the efficiency of Boolean satisfiability (SAT) solving. The proposed system, \emph{LangSAT}, bridges the gap between natural language inputs and propositional logic by converting English descriptions into  Conjunctive Normal Form (CNF) expressions and solving them using an RL-enhanced CDCL SAT solver. Unlike existing SAT-solving platforms that require CNF as input, LangSAT enables users to input standard English descriptions, making SAT-solving more accessible.

The framework comprises two key components: \emph{Lang2Logic}, which translates English sentences into CNF expressions, and \emph{SmartSAT}, an RL-based SAT solver. SmartSAT encodes clause-variable relationships as structured graph representations and extracts global features specific to the SAT problem. This implementation provides the RL agent with deeper contextual information, enabling SAT problems to be solved more efficiently.

Lang2Logic was evaluated on diverse natural language inputs, processing descriptions up to 450 words. The generated CNFs were solved by SmartSAT, which demonstrated comparable performance to traditional CDCL heuristics with respect to solving time. The combined LangSAT framework offers a more accessible and scalable solution for SAT-solving tasks across reasoning, formal verification, and debugging.

\end{abstract}

\section{Introduction}
The Boolean satisfiability (SAT) problem is a classic NP-complete~\cite{cook2021complexity} computer science problem regarding whether a given Boolean formula can be satisfied by assigning truth values to its variables. SAT problems are applied to model many problems, including hardware and software verification, cryptography, and equivalence checking \cite{selsam2019learning} \cite{marques2021conflict}.

One of the most prominent fields in SAT-solving is researching new techniques on existing algorithms, such as Conflict-Driven Clause Learning (CDCL). CDCL is a complete and efficient algorithm that has been the basis for many modern SAT solvers~\cite{fu2004zchaff,een2005minisat,soos2016cryptominisat}. CDCL improves the Davis-Putnam-Logemann-Loveland (DPLL) algorithm~\cite{davis1962}, a backtracking search algorithm that uses unit propagation and conflict analysis to solve SAT instances. DPLL is a sound and complete algorithm but inefficient for large instances due to its exponential time complexity. While still exponential in time complexity, CDCL is more efficient than DPLL in practice due to its ability to learn from conflicts and make informed decisions about variable assignments~\cite{marques2021conflict}.

Simultaneous developments in reinforcement learning (RL) and natural language processing (NLP) have opened up new horizons in human-computer interaction. Developments in NLP technology have positively impacted translation, conversational AI, and text-to-speech. Likewise, developments in the area of RL have accelerated innovation in robotics, autonomous vehicles, and the healthcare industry. However, both these areas have not yet been sufficiently explored in mathematical and logic-based domains, particularly SAT solving.

Most SAT solvers that use traditional algorithms require input to be in the CNF or simply logic statements and formally incompatible with the natural language format. This becomes a strong barrier, making problems in real applications not easily reformulated as an SAT instance. Thus, not having an accessible direct and computerized pipeline connecting natural language representations and formal logic hampers the broader use of SAT-solver techniques.

In order to address this gap, we propose a new methodology that integrates NLP and RL-based SAT solving. While NLP techniques facilitate the conversion of natural language statements to CNF, traditional CDCL solvers are perhaps not the optimal approach to proving and solving such translated instances. CNFs based on natural language inputs will generally consist of diverse structure patterns due to linguistic ambiguity and hence are unlike hand-crafted or algorithm-generated SAT instances. An RL-strengthened CDCL solver such as SmartSAT dynamically updates its decision heuristics depending on the structure of the problem so that it behaves uniformly with different CNF formulations. By leveraging reinforcement learning, the solver optimizes decision-making during the solving process, making it better suited for handling the variability introduced by NLP-generated CNFs.

This work, therefore, introduces a complete pipeline uniting natural language and SAT-solving procedures. By fusing NLP for CNF conversion and RL-strengthened CDCL solvers in verification, we demonstrate an accessible, efficient, and adaptable approach to SAT solving. This approach leads to possibilities in logic-based technology applications in everyday automated reasoning, software verification, and formal logic processing.

\section{Related Work}
\subsection{Overview of Methods for CNF Conversion in SAT Solvers}
Converting logical expressions into CNF is a crucial preprocessing step in SAT solving. The Tseytin transformation is one of the most widely used techniques, which converts arbitrary Boolean formulas into equisatisfiable CNF formulas by introducing auxiliary variables. This method ensures that the size of the CNF formula remains linear relative to the original formula~\cite{tseitin1983complexity}. However, other methods, such as multiplying out, generate CNF by directly applying logical equivalences, though they may result in exponentially larger formulas, which can be inefficient for SAT solvers. Moreover, the translation of natural language sentences into CNF expressions has attracted attention as an area of research. While no direct platform fully automates this process from English sentences to CNF, various tools have emerged to aid the conversion. NLP libraries, such as Natural Language Toolkit (NLTK), allow for parsing sentences into formal representations like Context-Free Grammars (CFGs), which can then be transformed into CNF~\cite{bird2006nltk}. Additionally, online grammar conversion tools provide functionalities to convert CFGs to CNF, though they require input in a specific formal grammar form~\cite{CyberZHG2020}. These combined approaches highlight the complexity and ongoing efforts in developing tools for efficient CNF generation for SAT solvers.
\subsection{Boolean Satisfiability Problem}
In the SAT problem context, an instance is delineated as a structured sequence of variables organized in CNF. Within CNF, variables \(x_1, x_2, \dots, x_n\) are used to construct clauses \(C_1, C_2, \dots, C_m\). A clause is composed of literals, which are variables or their negations, connected by logical ORs (\(\vee\)), and a CNF formula is composed of clauses connected by logical ANDs (\(\wedge\)). 

\noindent An example of a CNF formula with two clauses is:
\[
(x_1 \vee \neg x_2) \wedge (\neg x_1 \vee x_3).
\]
If a set of variable assignments can be made such that the CNF evaluates to true, it is satisfiable (SAT). Otherwise, if no satisfying assignment is possible, it is unsatisfiable (UNSAT)~\cite{davis1962}.
\subsection {Conflict-Driven Clause Learning}
Conflict-driven clause Learning has been a fundamental technique used in modern SAT solvers since their conception in the mid-90s. CDCL works by looping through a series of subroutines: Unit Propagation, Pick Branching Variable, Conflict Analysis, and Backtracking \cite{marques2021conflict}. Unit Propagation is a process that systematically deduces the truth values of variables within the propositional formula based on the presence of unit clauses. A unit clause is a clause containing only one literal, leaving no choice in the assignment of truth values to variables within that clause. Therefore, upon encountering a unit clause, the solver must assign a truth value to the corresponding variable to satisfy the clause. Unit propagation also applies to clauses that are partially assigned, as whenever all literals are assigned false except for one remaining, unit propagation forces that literal to be assigned true to make the clause true.

\noindent The Pick Branching Variable subroutine is utilized for the decision heuristic of CDCL. After rounds of Unit Propagation, there is a point in CDCL where there is no deducible best choice for the next variable to assign. Different CDCL solver adaptations implement the decision step in various ways, though a standard approach is to maintain a variable assignment stack to guide the following decision logic. Conflict Analysis is a subroutine that learns a new clause from the most recent conflict and computes a backtracking decision level. Identifying conflict nodes-variables and their associated decision levels constitutes a key aspect of conflict analysis. These nodes represent the trail of assignments and decisions leading to the conflict. The conflict nodes are determined through a process of backtracking from the conflicting clause to the decision level where the conflict originated. The newly learned clause is added to the clause database to help comprehend past decision-making mistakes and assist in future variable assignments. Finally, Backtrack is used in conjunction with Conflict Analysis to unassign variables until a backtrack level is reached. The main loop of CDCL concludes when all clauses evaluate to True, indicating that the problem is satisfiable or SAT, where a specific combination of variable assignments allows the problem to be solved \cite{farooque2024neurodual}.
\noindent As for our specific implementation of CDCL, we utilize our reinforcement learning agent as the decision heuristic.

\section{Methodology}
The LangSAT framework consists of a sequence of steps, each tasked with ensuring the correct and effective translation of natural language statements into a CNF form suitable for a computer and solving the ensuing problem with an RL-tuned CDCL solver. LangSAT is divided into two broad components:
\begin{itemize}
    \item \textbf{Lang2Logic} – for translating from natural language to CNF.
    \item \textbf{SmartSAT} – RL-driven enhancement to CDCL SAT solvers.
\end{itemize}

\noindent Since SAT solvers require input in a rigid CNF format, the first task of LangSAT is converting English sentences into a formal CNF representation. It accomplishes this transformation in three stages:

\begin{enumerate}
    \item \textbf{Natural Language to Logical Expression:} The given English input is parsed by a large language model (LLM) to generate a formal logical expression. This step ensures that significant logical relationships are preserved in a form interpretable by subsequent processing steps.
    \item \textbf{Logical Expression to CNF Conversion:} The resulting logical expression is converted to CNF using custom grammar-based parsing and symbolic computation libraries.
    \item \textbf{CNF Simplification:} The output CNF is further refined by eliminating redundant clauses and applying a minimal representation, improving solver efficiency.
\end{enumerate}

\noindent This structured pipeline enables an end-to-end conversion from unstructured natural language to a canonical SAT problem representation, making SAT solvers accessible for real-world problems requiring human-readable inputs.
Once the CNF representation is acquired, it is fed into \textbf{SmartSAT}, a reinforcement learning-based SAT solver that enhances traditional CDCL solving through dynamic heuristic selection based on learned experiences. Instead of utilizing pre-determined and non-deep learning heuristics like Variable State Independent Decaying Sum (VSIDS) \cite{liang2015understanding}, SmartSAT employs reinforcement learning to dynamically optimize variable selection. The RL agent is trained to maximize SAT-solving efficiency using an observation space that encodes:

\begin{itemize}
    \item \textbf{Clause-variable relationships} in a structured graph representation.
    \item \textbf{Global problem features} extracted from CNF formulations.
    \item \textbf{Current clause evaluations and variable assignments.}
\end{itemize}

By leveraging adaptive decision-making, SmartSAT generalizes across different SAT instances, offering a more flexible and responsive approach compared to static heuristic methods.

\subsection{Lang2Logic: Translating Natural Language to CNF}
The Lang2Logic framework aims to bridge the gap between natural language understanding and computational logic by converting English text into CNF. This transformation is essential for SAT solvers, as they operate on CNF representations. Lang2Logic consists of three core components: (1) Natural Language to Logical Expression, (2) Logical Expression to CNF Conversion, and (3) CNF Simplification. Each component is critical in ensuring an accurate and efficient transformation from natural language to a machine-readable CNF representation.

\begin{figure}[h]
    \centering
    \includegraphics[scale=0.23]{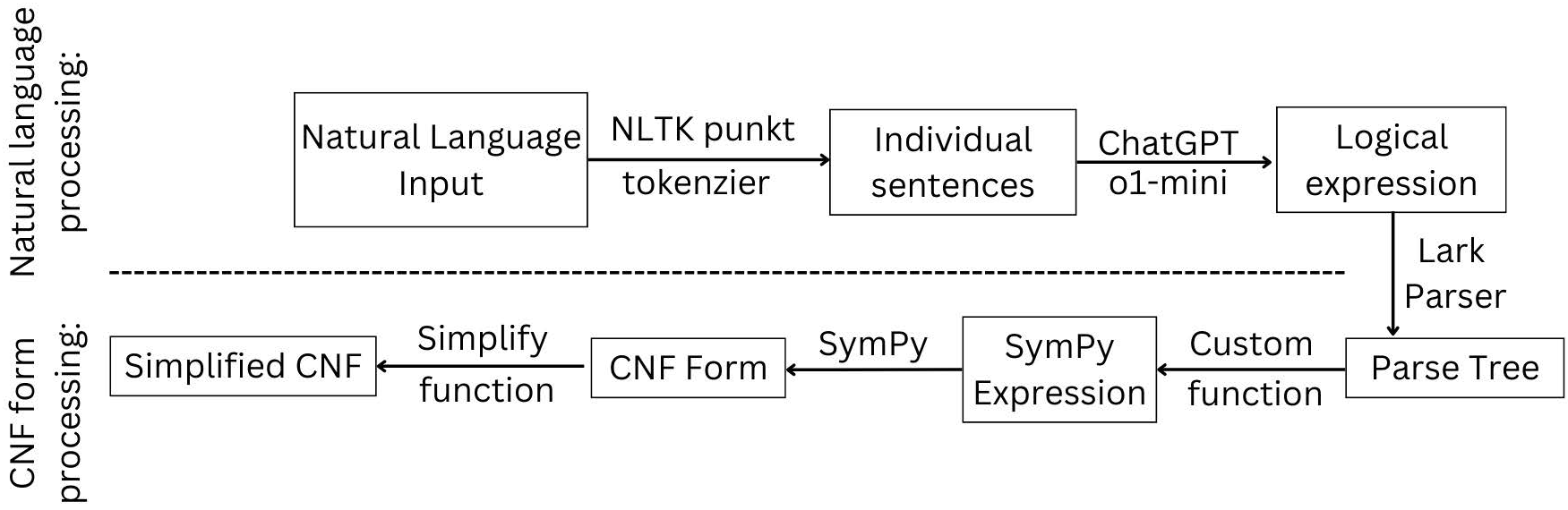}
    \caption{Lang2Logic framework: Flow from natural language input to CNF output.}
    \label{fig:lang2logic}
\end{figure}

Figure~\ref{fig:lang2logic} illustrates the overall flow of the Lang2Logic framework, depicting the sequential stages from natural language input to CNF output.

\subsubsection{Natural Language to Logical Expression}
The first stage utilizes the ChatGPT o1-mini model~\cite{ChatGPT} API to translate natural language input into a logical expression. The prompt for the o1-mini model was carefully designed to:
\begin{itemize}
    \item Recognize repeated variables across different clauses.
    \item Output fixed logical expressions in a structured format compatible with downstream processing.
    \item Ensure that the output aligns with the required input format for the \texttt{Sympy} converter~\cite{meurer2017sympy}.
\end{itemize}

The use of the ChatGPT API removes the need for hard-coded grammar definitions, simplifying the conversion process. However, the approach has limitations, such as higher costs and potential instability in performance as the input size increases. To mitigate these issues, the NLTK~\cite{bird2009natural} was employed to preprocess the input.
Using NLTK's Punkt tokenizer, the input text is split into individual sentences. This ensures that the o1-mini model processes each sentence independently, preventing errors such as sentence merging or skipping during translation. By working sentence-by-sentence, the o1-mini model can produce more accurate logical expressions. Example~\ref{ex:engtoLog} demonstrates the translation of an English sentence into an equivalent propositional logical expression.

\begin{example}
\label{ex:engtoLog}
\begin{itemize}
\item \textbf{Input Sentence:} \emph{"The circus has a Ferris wheel or a rollercoaster."}
\item \textbf{Logical Expression:} $\text{Or}(P, Q)$, where $P$ represents \emph{"The circus has a ferris wheel"} and $Q$ represents \emph{"The circus has a rollercoaster."}
\end{itemize}
\end{example}

\subsubsection{Logical Expression to CNF Conversion}
Once the natural language input has been converted to logical expressions, the next stage involves transforming these expressions into CNF. This is accomplished using a Lark parser~\cite{larkparser} and a predefined grammar. Each logical expression is processed as follows:
\begin{enumerate}
    \item \textbf{Parsing:} Each line is passed through the Lark parser to generate a parse tree. The parse tree organizes variables and logical operators into a structured hierarchy.
    \item \textbf{Sympy Conversion:} A custom function reads the parse tree and translates it into a \texttt{Sympy} expression. \texttt{Sympy} then handles the conversion of these expressions into CNF form.
\end{enumerate}


\begin{example}
\label{ex:logtocnf}
Let $P$ represent \emph{"The circus does not have a carousel"}, $Q$ represents \emph{"The circus has a ferris wheel"}, and $R$ represents \emph{"The circus has a rollercoaster"} and let the logical expression be
$\text{And}(\text{Not}(P), \text{Or}(Q, R))$.

The Lark parser generates a structured hierarchy where:
\begin{itemize}
\item $Not(P)$ is identified as a negation operator applied to $P$.
\item $Or(Q, R)$ is identified as a disjunction between $Q$ and $R$.
\item $And(\cdot)$ connects these components into a single logical expression.
  \end{itemize}
\textbf{CNF Output} $\sim P \land (Q \lor R)$ is converted to CNF as $(\sim P \lor Q) \land (\sim P \lor R)$.
\end{example}

\subsubsection{CNF Simplification}
The final module simplifies the CNF expressions for computational efficiency. This step leverages SymPy's \texttt{simplify\_logic} method, which reduces redundant clauses and literals while preserving logical equivalence and minimizing complexity. Simplifying the CNF formula ensures the resulting expression is computationally optimal for downstream tasks, such as SAT solving.

\begin{example}
Let \textbf{input:} $P \land (\neg Q \lor R) \land (\neg Q \lor R)$.\\
\textbf{Simplified output} is $P \land (\neg Q \lor R)$
\end{example}

The simplification process eliminates redundant clauses, retaining only the unique logical components of the formula. This ensures the CNF is compact while preserving its satisfiability properties, enabling more efficient processing by SAT solvers. 

\subsection{SmartSAT: RL-Driven Heuristic Optimization for SAT Solvers}
\vspace{0.1cm}
\subsubsection{Reinforcement Learning}
RL is a foundational approach in artificial intelligence. It enables systems (agents) to learn optimal behavior through interactions with their environment. In RL, the agent performs actions in a given state, observes the outcomes, and receives feedback in the form of rewards or penalties. Over time, the agent adjusts its decision-making policy to maximize cumulative rewards. 

Supervised learning methods, in particular variations of graph neural networks (GNNs), have been explored to facilitate decision heuristics in the CDCL-solving process.
In NeuroSAT, a GNN is essentially employed to predict satisfying assignments to an SAT problem by iteratively refining node representations and outputting final node values representing probabilities corresponding to variable assignments \cite{selsam2019learning}. Similarly, NeuroDual sought to use a Graph Attention Network (GAT), a specific form of a GNN, as a CDCL-solving heuristic \cite{farooque2024neurodual}. Both supervised learning approaches produced noticeable improvements. However, RL was chosen for the LangSAT framework due to several distinct advantages in solving Boolean satisfiability problems:
\begin{itemize}
    \item \textbf{Lack of Labeled Data:} Supervised learning requires extensive labeled datasets, which are scarce for SAT-solving problems. RL bypasses this requirement by learning through interactions with its environment rather than relying solely on pre-labeled training sets.
    \item \textbf{Sequential Nature of SAT Solving:} SAT problems involve multiple interdependent decision-making steps. Supervised learning methods struggle with sequential dependencies, whereas RL excels by dynamically adapting decisions based on the current state.
    \item \textbf{Adaptability to Problem Variability:} SAT problems exhibit diverse structures and complexities. RL adapts to these variations by dynamically tailoring its strategies to the problem at hand, whereas supervised methods often fail to generalize across diverse problem instances.
\end{itemize}

\noindent LangSAT's reinforcement learning policy leverages Proximal Policy Optimization (PPO), a robust and efficient policy gradient method. PPO optimizes a probabilistic policy, linking observations to actions while ensuring stable training by constraining policy updates within a safe range \cite{schulman2017proximal}.
\subsubsection{Observation Space}
The RL agent's observation space is designed to provide critical problem-specific and global information, prompting effective decision-making. It includes:
\begin{itemize}
    \item \textbf{Current Variable Assignments:} Tracks assigned and unassigned variables. Variables assigned True are represented as $+1$, False as $-1$, and unassigned as $0$.
    \item \textbf{Current Clause Evaluations:} Indicates whether clauses evaluate to True ($+1$), False ($-1$), or are unevaluated ($0$). This guides the agent toward satisfying all clauses.
    \item \textbf{Clause-Variable Graph Representations:} The SAT problem is represented as a \textit{bipartite graph} where:
    \begin{itemize}
        \item Variables and clauses are modeled as distinct nodes.
        \item Edges exist between a variable node and a clause node if the variable appears in that clause.
        \item Positive and negative literals are treated distinctly to preserve clause polarity.
    \end{itemize}
    This graph-based representation allows the RL agent to learn structural dependencies between clauses and variables dynamically, similar to NeuroDual \cite{farooque2024neurodual}, which utilized a GAT for clause-variable encoding. Instead of applying a purely supervised GAT approach, SmartSAT integrates reinforcement learning with this clause-variable graph structure, enabling adaptive heuristic learning.
    \item \textbf{Global SAT Problem Features:} Extracted features provide overarching context applicable to any SAT problem, complementing local graph information.
\end{itemize}

\subsubsection{Action Space and Rewards}
The RL agent's action space comprises 40 actions, mapping each variable to a Boolean assignment (True or False). During training, the agent receives:
\begin{itemize}
    \item \textbf{Reward:} $+1$ for each satisfied clause.
    \item \textbf{Penalty:} $-1$ for each unsatisfied clause.
\end{itemize}

\noindent The maximum achievable reward is $91$, corresponding to all clauses being satisfied. Through the PPO policy, the agent iteratively refines its strategy to maximize cumulative rewards, evolving into an efficient decision heuristic for SAT-solving.

\subsubsection{Feature Selection}
The core of NeuroDual stems from its multifaceted feature selection process. We select a total of 48 features, aggregating various information on the model instance with which we are working. These features are shared between all nodes, as they are global CNF information obtained from the SATfeatPy library \cite{provan2022satfeatpy}. SATfeatPy offers a rich array of different feature embeddings for a DIMACS-CNF clause, assembling features used from SATzilla \cite{xu2008satzilla}, \cite{ansotegui2017structure}, and \cite{alfonso2014new}. We utilized SATfeatPy's method to obtain 48 features from SATzilla base \cite{xu2008satzilla}, a highly accurate portfolio-based SAT solver. By leveraging the same core features from SATzilla, our model has sufficient global information to understand the context of the problem.

\subsubsection{ Dataset}
DIMACS (The Center for Discrete Mathematics and Theoretical Computer Science) has developed a standardized format for encoding CNF files, deemed DIMACS-CNF. SATLIB \cite{hoos2000satlib} offers a benchmark of CNF instances for the training and evaluating SAT solver models.
Many available DIMACS-CNF files can be obtained through SATLIB. Because of the standardized format of DIMACS-CNF, making a parser for a wide array of clauses is trivial. We adopted the use of uf20-91, which is the shorthand file name for satisfiable instances of clauses comprised of 20 variables and 91 clauses \cite{farooque2024neurodual}.

\subsubsection{Agent Training and Tuning}

Every training step of SmartSAT occurs at every decision point in our RL-integrated CDCL solver. By training our agent at each decision point, it can be penalized and rewarded more frequently, thus equipping it to learn how to solve SAT problems more effectively.
Our training process began by dividing the \texttt{uf20-91} dataset into a training and testing split of \(80\%\) and \(20\%\), respectively, resulting in a training set of 800 CNF files and a testing set of 200 CNF files. To optimize our model, we utilized the PPO policy gradient method previously discussed. Additionally, we applied a standard learning rate of 0.0002, as we found through trials that it allowed our model to learn the dynamics of various SAT problems at a sufficient rate. SmartSAT then trained for one full epoch on its training set, which corresponded to roughly 100,000 training steps.

\subsubsection{SmartSAT Algorithm}
The SmartSAT Algorithm starts by receiving the converted CNF form of the SAT problem from Lang2Logic. The subsequent steps are carried out by a reinforcement learning-based CDCL solver. The solver follows a series of iterative steps:
\begin{itemize}
   \item The RL model analyzes its observation space, which consists of:
    \begin{itemize}
        \item Variable assignments
        \item Clause evaluations
        \item A structured clause-variable representation graph
        \item Global SAT features
    \end{itemize}  
    \item Based on its trained PPO policy, the RL agent selects the following variable to assign and determines its Boolean value.
    
    \item The CDCL solver continues functioning normally, following Boolean Constraint Propagation (BCP), until a conflict is encountered or the problem is classified as satisfiable or unsatisfiable.
    
    \item If a conflict occurs, CDCL performs backtracking to obtain a new learned clause and updates the clause database to refine future decisions.
    
    \item The RL agent makes another decision, which repeats iteratively until the problem is deemed SAT or UNSAT.
\end{itemize}

\noindent The algorithm is detailed in Algorithm~\ref{alg:SmartSAT}. 
\begin{algorithm}
\caption{SmartSAT Algorithm}
\label{alg:SmartSAT}
\textbf{Input:} Converted CNF form of the SAT problem from Lang2Logic \\
\textbf{Output:} SAT/UNSAT classification
\begin{algorithmic}[1]
    
    \State \textbf{CDCL solver execution begins}
    \While{CDCL solver is running}
    
        \Statex // Extract RL observation space at each step
        \State \textit{obsSpace} $\gets$ extractObservationSpace()

        \Statex // RL Agent Decision Step
        \State $\textit{assignment} \gets$ RL\_agent\_decision(\textit{obsSpace})

        \Statex // Assign Variable: Apply RL decision
        \State ApplyAssignment(\textit{assignment})

        \Statex // Perform Boolean Constraint Propagation (BCP)
        \State propagateConstraints()

        \If{SAT solution is found}
            \State \Return SAT with satisfying assignment
        \ElsIf{Conflict is encountered}
        
            \Statex // Conflict Resolution and Backtracking
            \State learnNewClause()
            \State updateClauseDatabase()
        \EndIf
    \EndWhile

    \State \Return UNSAT
\end{algorithmic}
\end{algorithm}

\subsubsection{Comparison to Baseline CDCL Model}
Our primary method for comparing the effectiveness of SmartSAT to Baseline CDCL solvers is through total solving time. Specifically, we run both SmartSAT and a traditional
CDCL solver utilizing the VSIDS heuristic under identical test sets and conditions: varying problem structures and complexities, and the computing environments and solving parameters are the same. This comparison strategy provides insight into the efficiency of variable selection for our RL agent and the decision-making strategy used to tackle the SAT problem. Figure~\ref{fig:smartsat-diagram} below illustrates the two SAT-solving architectures being compared. The objective is for SmartSAT to achieve higher performance over traditional CDCL through the addition of RL-related architectural blocks. The RL agent should be able to learn an effective PPO policy, enabling it to make more efficient decisions than VSIDS after CDCL backtracking is performed. 
\begin{figure}
    \centering
    \includegraphics[scale=0.37]{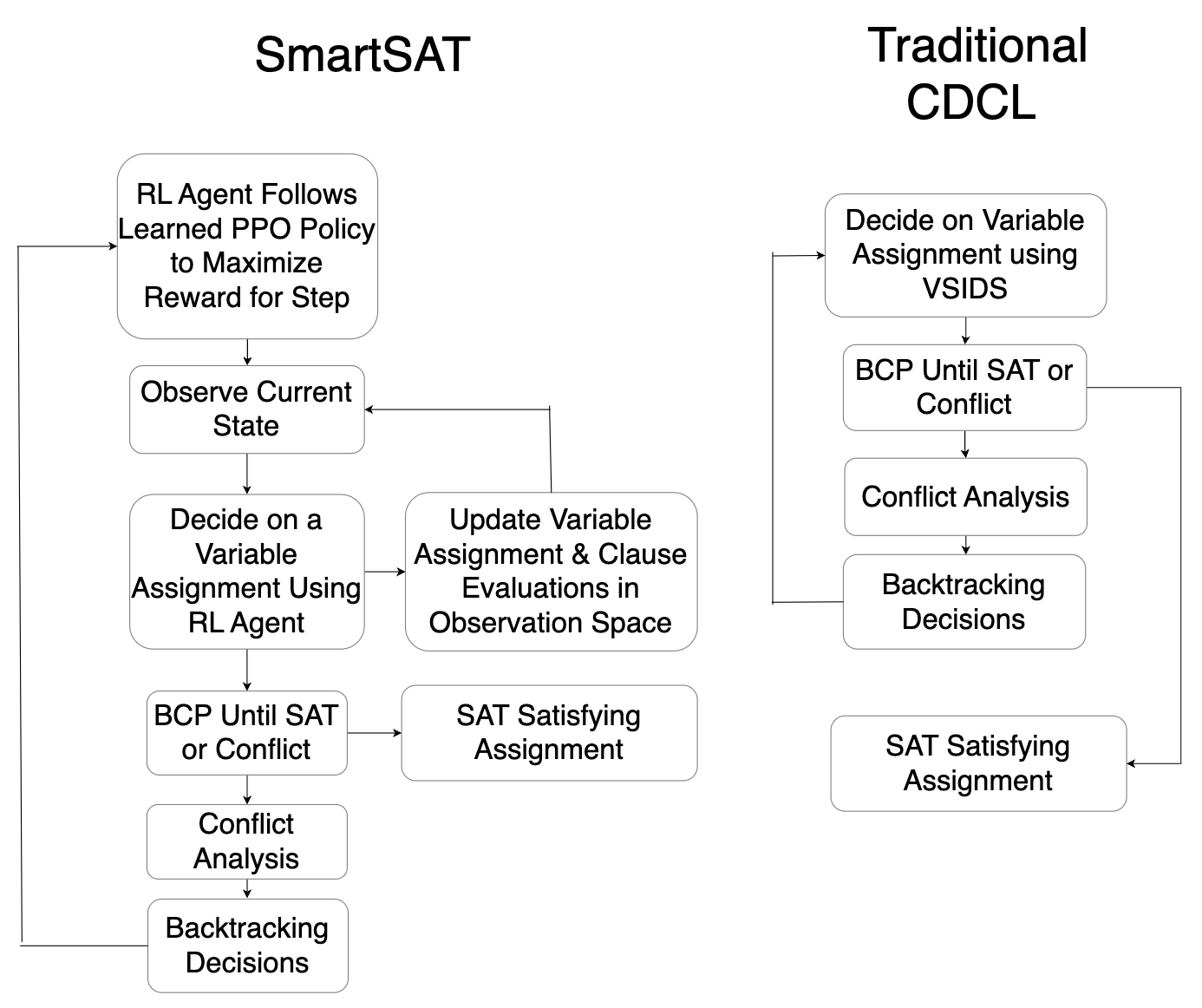}
    \caption{Architecture comparison of SmartSAT vs. Traditional CDCL}
    \label{fig:smartsat-diagram}
\end{figure}

\section{Results}
\begin{figure}[h]
    \centering
    \includegraphics[scale=0.06]{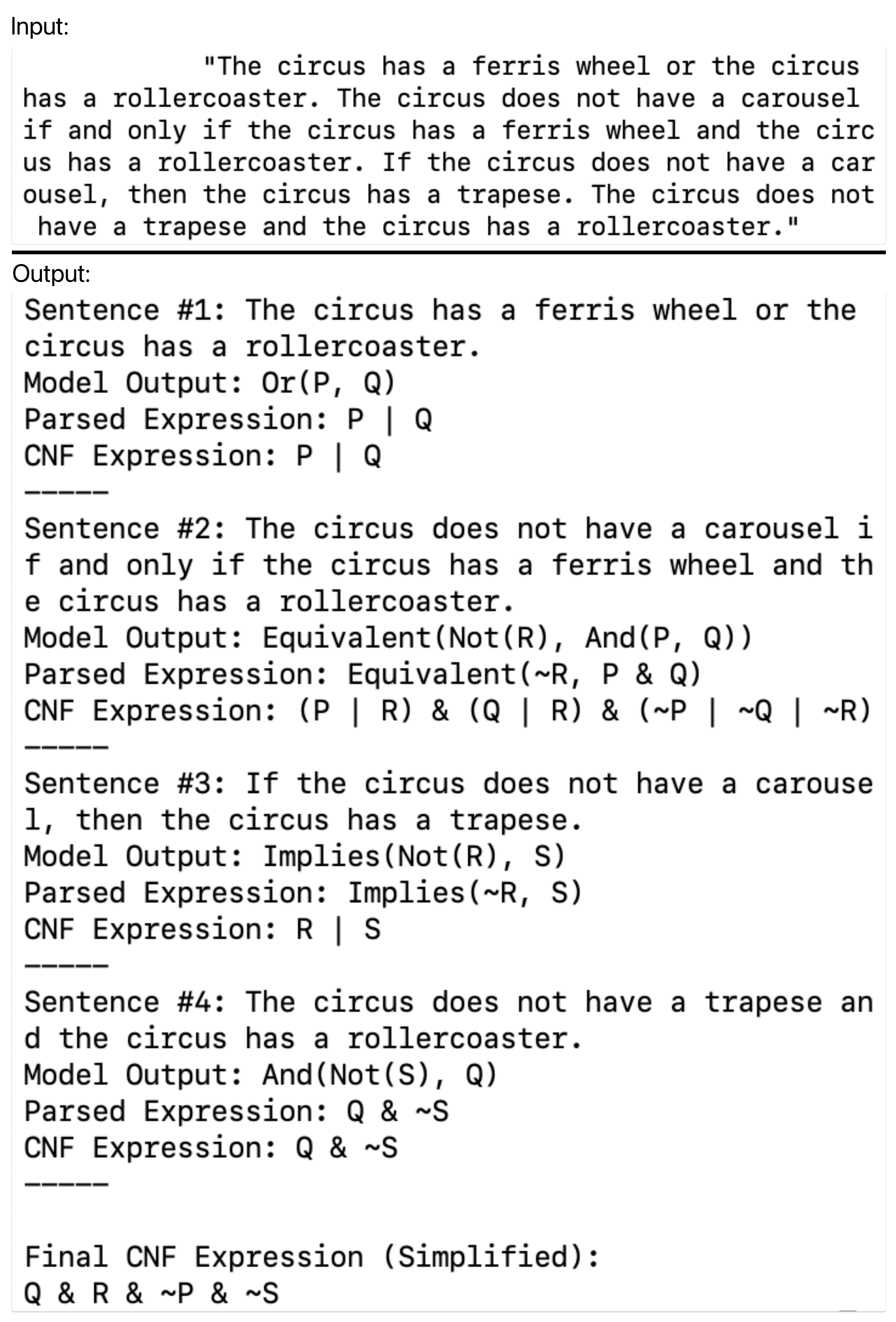} 
    \caption{Lang2Logic test case with four clauses and four variables}
    \label{fig:rl-agent-diagram}
\end{figure}

Lang2Logic digests paragraphs into individual sentences, translates each English sentence into a logical expression, and then converts it to CNF form. It also simplifies the final CNF expression for downstream usage. Figure~\ref{fig:rl-agent-diagram} shows the input paragraph at the top. This paragraph is analyzed by Lang2Logic and converted into four clauses with four variables in CNF form. Lastly, Lang2Logic performs simplifications on this CNF form. The final corresponding output is displayed on the bottom portion of the figure.

\begin{figure}
    \centering
    \includegraphics[scale=0.25]{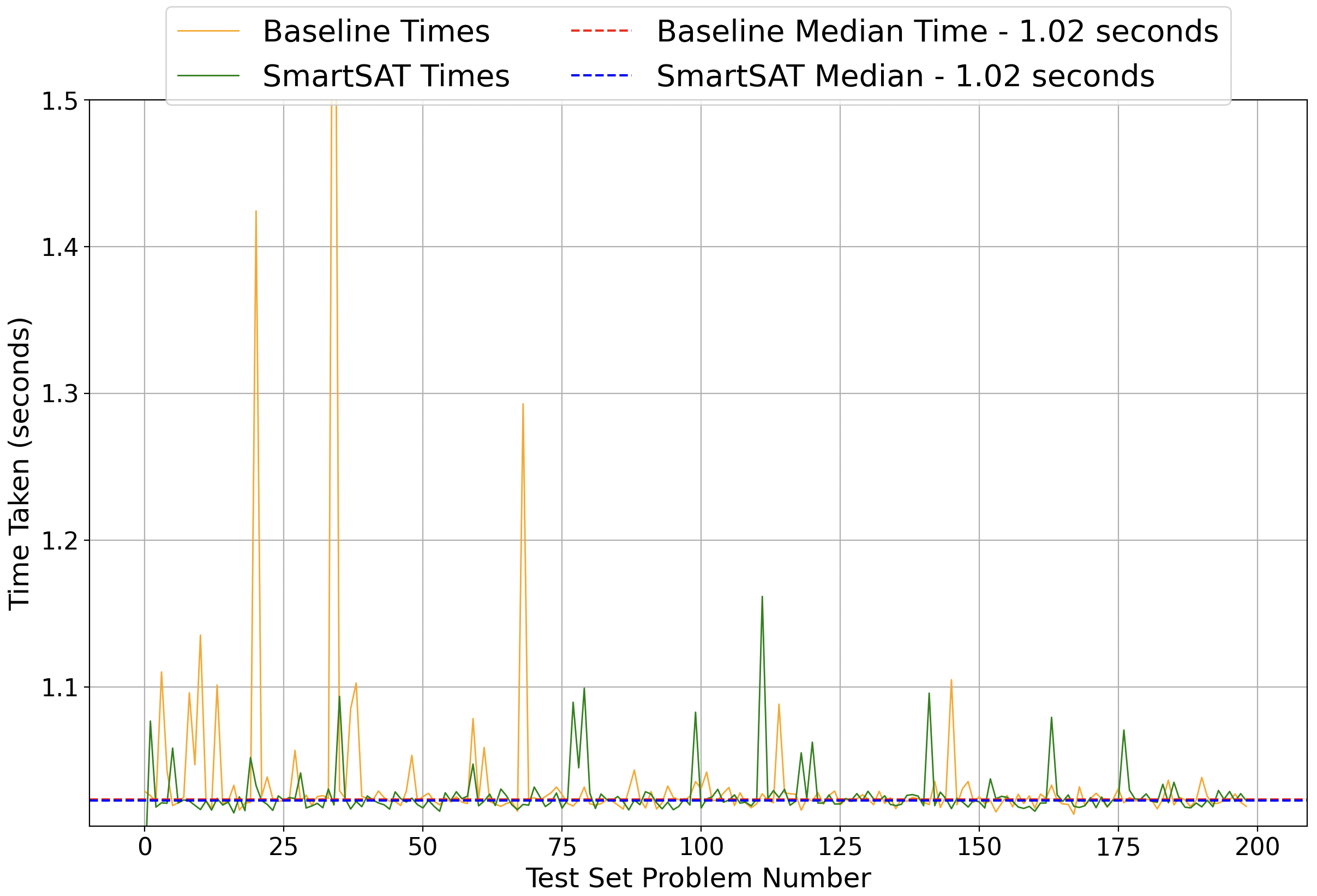} 
    \caption{Comparison of Baseline Times and SmartSAT Times for \texttt{uf20-91} Test Set}
    \label{fig:rl-agent-diagramT}
\end{figure}

Figure~\ref{fig:rl-agent-diagramT} showcases the comparison of the total time taken to solve each individual CNF file in our test set for a baseline CDCL solver using the VSIDS heuristic and LangSAT. Both models solved each problem in the \texttt{uf20-91} testing set within a range of 1.01 to 1.05 seconds, with most problems clustering at 1.01 to 1.02, as depicted by the graph. In addition, the visual trend suggested that SmartSAT demonstrates slightly higher solving time performance over the baseline model, achieving more consistent results with fewer outliers. 

Both the dashed lines represent the \texttt{uf20-91} test set median solving time for each model. As shown, the median solving times for both the Baseline CDCL model and SmartSAT are roughly tied at 1.02 seconds. In addition, the total problems that SmartSAT demonstrated faster solving times versus the baseline model was calculated at roughly 53\%, indicating that SmartSAT marginally surpasses the performance level of the traditional CDCL solver.

The results further provide evidence that SmartSAT's approach is adaptable and efficient in handling various SAT problem structures. Although both models attained nearly identical median times, SmartSAT performed more consistently across a broader range of CNF structures, as illustrated by the more frequent spikes in solving time for the CDCL solver. Baseline CDCL likely performed poorly in some instances due to its static heuristic approach. Conversely, SmartSAT was able to generalize across a wide range of SAT problems due to the RL-based decision-making. This ability to produce consistent results makes RL-integrated SAT solvers very promising, particularly in situations where problem structures differ significantly.
\section{Conclusion}
Our work demonstrates significant success by developing LangSAT, a modern, innovative SAT-solving framework combining the power of a natural language conversion with RL-based CDCL solvers that generated comparable results to traditional SAT-solving methods. The introduction of LangSAT marks a step forward in the accessibility and efficiency of SAT-solving frameworks.

Lang2Logic can accept natural language as input, which brings advantages to the diversity of inputs for the SAT-solving framework. It breaks the limitation of only the CNF form serving as input to SAT solvers, multiplying the ability and usability of SAT solvers from traditional logical problems to language processing problems. With low costs involving only the ChatGPT API, it delivers reliable translation from Natural Language to CNF form. Lang2Logic is a reliable bridge to SmartSAT, opening more possibilities for the types of input it can process.

Analyzing the SmartSAT component of LangSAT in particular, our work found that RL-based decision heuristics integrated with traditional CDCL algorithms can learn key pieces of structural information in the context of SAT problems. Our results prove that LangSAT performs better than a standard CDCL model on approximately half of all issues in the \texttt{uf20-91} dataset. SmartSAT can perform comparably well to traditional CDCL solvers due to the encoding of clause-variable relationships as structured graph representations and through the extraction of global features specific to the SAT problem.

With further improvements, LangSAT can extract the proper CNF form from larger descriptions of text and achieve a shorter solving time for most instances compared to traditional state-of-the-art CDCL solvers. To further increase input diversity, machine code is the next direction we are pursuing. As architectural pieces of software building, code represents the fundamental units that carry basic information for computer communication. Lang2Logic will adopt code as input to broaden SmartSAT's usability further and provide more flexibility in input selection. As for the SmartSAT component of LangSAT, we plan to make a few refinements. Firstly, we aim to provide our RL agent with additional global features for our SAT instances so that it can decipher more patterns from each problem. Another part of our future work for SmartSAT consists of generating total time comparisons across various SAT problem datasets. This would allow us to analyze whether or not our RL-based heuristic for SmartSAT performs better for certain types of SAT problems and worse for others. 


%
%
%
 \bibliographystyle{splncs04}
 \bibliography{cade}
\end{document}